\documentclass[letterpaper]{article} 
\usepackage{aaai2026}  
\usepackage{times}  
\usepackage{helvet}  
\usepackage{courier}  
\usepackage[hyphens]{url}  
\usepackage{graphicx} 
\usepackage{natbib}  
\usepackage{caption} 
\usepackage{algorithm}
\usepackage{algorithmic}

\usepackage{amssymb} 
\usepackage{amsmath}

\usepackage{pifont}
\usepackage{xcolor}
\usepackage{multirow}
\usepackage{tcolorbox} 
\tcbuselibrary{skins,breakable}
\usepackage{parskip}   

\newcommand{\method}{\texttt{Note2Chat}}

\usepackage{newfloat}
\usepackage{listings}
\DeclareCaptionStyle{ruled}{labelfont=normalfont,labelsep=colon,strut=off} 
\floatstyle{ruled}
\newfloat{listing}{tb}{lst}{}
\floatname{listing}{Listing}
\title{\method{}: Improving LLMs for Multi-Turn Clinical History Taking \\ Using Medical Notes}
\author{
    Yang Zhou\textsuperscript{\rm 1}\equalcontrib\textsuperscript{\textdagger}, Zhenting Sheng\textsuperscript{\rm 2}\equalcontrib, 
    Mingrui Tan\textsuperscript{\rm 1}, Yuting Song\textsuperscript{\rm 1}, Jun Zhou\textsuperscript{\rm 1}, \\
    Yu Heng Kwan\textsuperscript{\rm 3,\rm 4}, Lian Leng Low\textsuperscript{\rm 3,\rm 4}, Yang Bai\textsuperscript{\rm 1}\thanks{Corresponding Authors.}, Yong Liu\textsuperscript{\rm 1} \\
}
\affiliations{
    \textsuperscript{\rm 1}Institute of High Performance Computing (IHPC), Agency for Science, Technology and Research (A*STAR), Singapore\\

    \textsuperscript{\rm 2}Nanyang Technological University \\
    \textsuperscript{\rm 3}National University of Singapore \\ \textsuperscript{\rm 4}Singapore General Hospital 

}

\begin{document}

\maketitle

\begin{abstract}
Effective clinical history taking is a foundational yet underexplored component of clinical reasoning. While large language models (LLMs) have shown promise on static benchmarks, they often fall short in dynamic, multi-turn diagnostic settings that require iterative questioning and hypothesis refinement. To address this gap, we propose \method{}, a note-driven framework that trains LLMs to conduct structured history taking and diagnosis by learning from widely available medical notes. Instead of relying on scarce and sensitive dialogue data, we convert real-world medical notes into high-quality doctor-patient dialogues using a decision tree-guided generation and refinement pipeline. We then propose a three-stage fine-tuning strategy combining supervised learning, simulated data augmentation, and preference learning. 
Furthermore, we propose a novel single-turn reasoning paradigm that reframes history taking as a sequence of single-turn reasoning problems.
This design enhances interpretability and enables local supervision, dynamic adaptation, and greater sample efficiency. 
Experimental results show that our method substantially improves clinical reasoning, achieving gains of +16.9 F1 and +21.0 Top-1 diagnostic accuracy over GPT-4o.
\end{abstract}

\begin{links}
    \link{Code}{https://github.com/zhentingsheng/Note2Chat}
\end{links}

\section{Introduction}
History taking and differential diagnosis are fundamental to clinical reasoning, forming the basis for understanding a patient's condition and directing subsequent diagnostic and therapeutic decisions. A thorough history typically encompasses the chief complaint, history of present illness, review of systems, and general medical and social background, all of which collectively inform the generation and refinement of a differential diagnosis, a ranked list of plausible conditions grounded in the patient’s symptoms and risk factors. Central to this process is \textit{multi-turn clinical history taking}, a dynamic, interactive dialogue in which clinicians iteratively ask targeted questions, interpret responses in context, and update diagnostic hypotheses step by step \cite{Henderson2012PatientHistory,GuyattEtAl2015}. This sequential reasoning demands broad medical knowledge, adaptability, and time—resources often constrained in high-volume care environments. Despite these challenges, history taking alone can lead to accurate diagnoses in a majority of cases \cite{Kuriakose2020, nierenberg2020using}, underscoring its diagnostic value. In this context, automated history-taking systems hold significant promise: by conducting structured, multi-turn interviews prior to clinical encounters, they can streamline information gathering, reduce cognitive load on clinicians, and enhance the completeness and consistency of diagnostic conversations, particularly in settings with limited medical personnel.

\begin{figure}[tbp]
  \centering
  \includegraphics[width=0.98\linewidth]{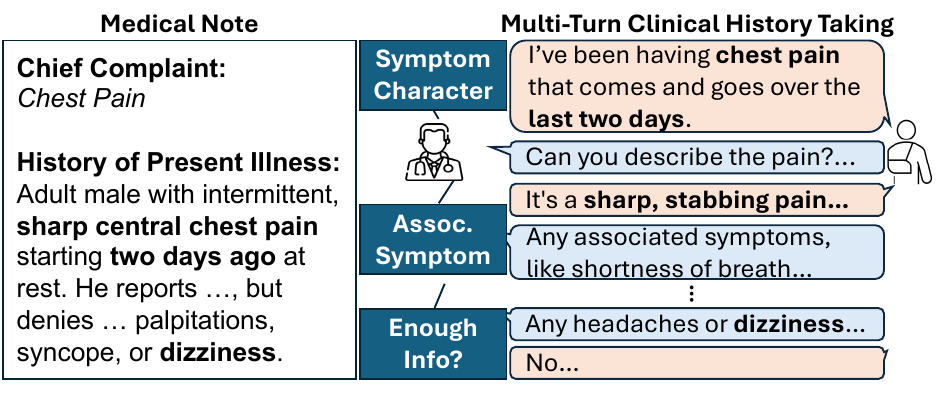}
  \caption{Multi-turn clinical history taking is the dynamic process of gathering information from a patient; structured medical notes are the organized product, synthesizing that narrative into a concise, standardized record.}
  \label{fig:intro}
\end{figure}

Large language models (LLMs) have demonstrated impressive performance across a range of medical tasks, including medical question answering, clinical note summarization, and care plan generation \cite{Cabral2024ClinicalReasoning,Goh2024LLMInfluence,McDuff2025AccurateDifferential,Nori2023GPT4Medical,Nori2023GeneralistOutcompete,Nori2024FromMedprompt}. However, these achievements are largely based on static, single-turn benchmarks, where models are provided with complete clinical vignettes and tasked with producing an answer without needing to interact or inquire further. Such settings fail to capture the sequential and exploratory nature of real-world diagnostic reasoning, which requires actively gathering missing information through dialogue. Recent benchmark studies have underscored this limitation, showing that when LLMs are evaluated in full diagnostic conversations, where they must initiate questions, adapt based on responses, and iteratively refine hypotheses, their diagnostic accuracy can drop significantly compared to single-turn tasks \cite{Johri2025EvaluationFramework,Liu2024AutomaticEvaluation,Li2024MediQ,Hager2024LimitationsLLMs,Schmidgall2024AgentClinic}. This performance gap indicates that, despite their medical knowledge, current LLMs lack the conversational competencies needed for effective multi-turn diagnostic reasoning. In particular, they often fail to generate focused follow-up questions or to prioritize clinically relevant details, constraining their usefulness in structured interviews \cite{Goh2024LLMInfluence,Nori2024FromMedprompt}. These findings point to the need for dynamic, interaction-oriented evaluations that better reflect the challenges of history taking and conversational diagnosis.

Recent efforts to enhance LLMs for clinical dialogue have explored self-play simulation, agent-based workflows, and reinforcement learning (RL) strategies. For example, AMIE \cite{tu2025towards} introduced a simulated diagnostic environment to improve history-taking dialogue, but relies on private datasets and models. DoctorAgent-RL \cite{feng2025doctoragentrl} has introduced RL fine-tuning to encourage LLMs to progressively refine diagnoses through proactive questioning. Agent-based methods \cite{nori2025sequential,gatto2025followupq,liu2025doccha,rose2025meddxagent} assign different roles to separate LLMs, yet typically use general-purpose models not tailored for clinical reasoning. RL-based approaches \cite{fansi2022ddxplus,sun2025improving} aim to improve diagnostic performance through fine-tuning, but often depend on rigid supervision or task-specific annotations, limiting adaptability. Despite their differences, most of these works prioritize final diagnosis accuracy and underemphasize the quality and completeness of history taking. They often overlook clinically important but non-diagnostic details, such as negative findings or symptom context, and are limited by the lack of large-scale, high-quality dialogue data. Given the unique challenges of history taking, including its exploratory nature and variation across clinical styles, there is a clear need for scalable and generalizable training paradigms. 

In this work, we address the challenges of clinical history taking by focusing on efficient information gathering rather than optimizing solely for diagnostic accuracy. 
Our objective is to enable the model to extract as many relevant findings as possible with minimal questioning, supporting concise and complete interviews grounded in clinical reasoning. 
To this end, we propose \method{}, a novel framework that leverages real-world medical notes, specifically the primary diagnosis and history of present illness (HPI), as a supervision signal. 
These notes capture clinician-curated summaries of symptom relevance, temporal progression, and diagnostic thinking, offering a rich and widely accessible resource for training. 
Compared to medical dialogues, clinical notes are significantly more available, as they are routinely documented for care delivery and are less restricted by privacy concerns.
Moreover, they require no additional manual annotation and can be easily adapted to local protocols and institutional practices.

Our framework consists of three core components designed to enhance LLMs for clinical history taking. 
First, we introduce a \textbf{note-to-dialogue generation pipeline} that converts discharge notes into clinically meaningful doctor-patient conversations using decision tree-guided prompts, followed by refinement to ensure realism and comprehensive coverage of key findings. 
Using this pipeline, we construct a dataset comprising 8,944 synthetic dialogues, 67,077 successful rollouts, and 11,403 preference pairs across 4,972 patients. 
Second, we propose a \textbf{three-stage fine-tuning strategy} that combines supervised training on note-guided dialogues, data augmentation through simulated interactions, and direct preference optimization (DPO) \cite{rafailov2023direct} to encourage concise and clinically effective conversations. 
Third, we introduce a \textbf{single-turn reasoning paradigm} that treats each dialogue turn as an independent decision step, enabling the model to make context-aware, interpretable actions guided by conversation history and reasoning plans. This design improves follow-up questioning, streamlines information gathering, and allows for explicit reward shaping based on information gain and diagnostic relevance. Our contributions include: 
\begin{itemize}
    \item We propose a note-driven dialogue generation and refinement pipeline to curate clinically grounded patient-doctor interactions, resulting in a history-taking dataset across 4,972 patients.
    \item We propose a three-stage fine-tuning strategy and a single-turn reasoning paradigm that enhance LLMs for concise, interpretable, and effective history taking.
    \item Using \method{}, our fine-tuned LLM achieves state-of-the-art performance in both information gathering and diagnosis, with relative gains of +57.53\% and +42.86\% over GPT-4o, respectively.
\end{itemize}

\section{Related Work}
\textbf{LLMs for medical applications:} 
LLMs have demonstrated impressive capabilities across a broad range of medical applications, including question answering, clinical summarization, and care planning \cite{Cabral2024ClinicalReasoning,Goh2024LLMInfluence,McDuff2025AccurateDifferential,Nori2023GPT4Medical,Nori2023GeneralistOutcompete,Nori2024FromMedprompt,achiam2023gpt,saab2024capabilities}.
Advanced reasoning models such as DeepSeek-R1 \cite{guo2025deepseek} and Gemini 2.5 \cite{comanici2025gemini} continue to raise the bar with stronger generalization and reasoning performance on medical benchmarks. 
Meanwhile, domain-specialized variants like BioMistral \cite{labrak2024biomistral}, HuatuoGPT-o1 \cite{chen2024huatuogpt}, Med-R1 \cite{lai2025med}, and MedGemma \cite{sellergren2025medgemma} further tailor LLMs to clinical contexts by incorporating structured medical knowledge and fine-tuning on healthcare-specific data. 
Despite these advances, most evaluations remain confined to static, single-turn settings in which models are presented with complete case information and asked to produce a response. 
Such paradigms overlook the inherently dynamic and sequential nature of real-world clinical reasoning—particularly in tasks like history taking and differential diagnosis, where success hinges on actively collecting missing information, asking follow-up questions, and reasoning under uncertainty. 
Recent studies have shown that even state-of-the-art models struggle in these interactive, multi-turn environments, with diagnostic performance dropping significantly when required to reason step-by-step without full context \cite{Johri2025EvaluationFramework,Liu2024AutomaticEvaluation,Li2024MediQ,Hager2024LimitationsLLMs,Schmidgall2024AgentClinic}. 
These limitations underscore the need for new training and evaluation frameworks that support proactive, reasoning-driven dialogue and reflect the complexities of real clinical workflows.

\textbf{LLMs for multi-turn clinical conversation:} 
Recent research has increasingly focused on enabling LLMs to participate in multi-turn clinical conversations, particularly for tasks such as history taking and differential diagnosis. 
Several benchmarks 
\cite{Johri2025EvaluationFramework,Li2024MediQ,Schmidgall2024AgentClinic,wang2025healthq,chandra2025reasoning,fan2024ai} have been introduced to evaluate LLMs in interactive medical settings, offering useful tools for assessing question-asking and reasoning abilities. 
Multi-agent frameworks that assign roles like history taker or diagnostician to separate LLMs aims to improve the workflow rather than LLMs' clinical reasoning ability. 
RL-based methods \cite{fansi2022ddxplus,sun2025improving} promote proactive questioning and diagnostic refinement, but depend on rigid, pre-defined state-action spaces that limit flexibility and generalizability. 
AMIE \cite{tu2025towards} marks a notable advancement by training models in a self-play diagnostic environment and extending to multimodal data \cite{saab2025advancing}, yet it relies on proprietary datasets and closed-source models, limiting reproducibility and broader adoption. 
To address data scarcity, prior curation efforts \cite{tu2025towards,fansi2022ddxplus,saley2024meditod,chen2025clinichat} generate training data from annotated dialogues, synthetic vignettes, or QA-style conversions, and typically use supervised fine-tuning (SFT) for training. 
DoctorAgent-RL \cite{feng2025doctoragentrl} further applies RL fine-tuning to enhance question generation but remains constrained by synthetic, limited training resources.

\section{\method{}}
We introduce \textbf{\method{}}, a generalizable LLM training framework for clinically grounded, proactive history taking. By leveraging medical notes as a natural and scalable supervision source, our approach shifts the focus from diagnosis to high-quality information gathering, better aligning with the exploratory nature of real-world clinical reasoning.

\textbf{Problem Setup.}
We formulate medical history taking as a \textit{partially observable sequential decision-making process}, where a doctor agent interacts with a simulated patient grounded in clinical notes. 
The objective of the doctor agent is to elicit clinically relevant findings through follow-up questions and ultimately produce a differential diagnosis.

Let \( \mathcal{P} \) denote the distribution over patient cases. Each case \( x \sim \mathcal{P} \) is defined as \( x = \{ \text{dx}, \mathcal{F}, \text{cc} \} \), where \( \text{dx} \) is the ground-truth diagnosis extracted from the note, \( \mathcal{F} = \{f_1, \ldots, f_n\} \) is the set of clinical findings extracted from the \textit{History of Present Illness (HPI)}, and \( \text{cc} \) is the chief complaint, serving as the initial observation.
At each turn \( t \), the doctor agent observes a state \( s_t = \{ \text{cc}, h_t \} \), where \( h_t = \left[ (q_1, r_1), (q_2, r_2), \ldots, (q_{t-1}, r_{t-1}) \right] \) is the accumulated dialogue history, where $q_t$ and $r_t$ are the doctor question and patient response, respectively.
The agent selects an action \( a_t \in \mathcal{A} = \mathcal{A}^{\text{ask}} \cup \mathcal{A}^{\text{diagnose}} \), either asking a follow-up question $q_t$ or issuing a diagnostic prediction.

If the agent selects a question-asking action \( a_t \in \mathcal{A}^{\text{ask}} \), a simulated patient grounded in \( x \) responds with \( r_t \sim p(r \mid q_t, x) \), and the dialogue history is then updated to include the new exchange, resulting in the next state \( s_{t+1} = \{ \text{cc}, h_{t+1} \} \), where \( h_{t+1} = h_t \cup \{(q_t, a_t)\} \). Alternatively, if \( a_t \in \mathcal{A}^{\text{diagnose}} \),  the interaction terminates and the doctor agent will predicta ranked list of \( K \) potential diagnoses \( \hat{\mathbf{y}}_t = [\hat{y}_t^{(1)}, \hat{y}_t^{(2)}, \ldots, \hat{y}_t^{(K)}] \subset \mathcal{Y} \), where \( \mathcal{Y} \) is the space of diagnostic labels.
Each interaction forms a trajectory \( \tau = \left( s_0, a_0, s_1, a_1, \ldots, s_T, a_T \right) \), ending when a diagnosis is made. The doctor agent learns a policy \( \pi_\theta(a_t \mid s_t) \), parameterized by \( \theta \), governs decision-making.
The learning objective is to train the policy $\pi_\theta$ to maximize the expected reward $R(h_T)$, which evaluates the informativeness and efficiency of the dialogue:
$\max_\theta \mathbb{E}_{x \sim \mathcal{P}, \pi_\theta}[R(\tau)]$.
This can be optimized via supervised fine-tuning (SFT), reinforcement fine-tuning, or preference-based fine-tuning guided by note-derived supervision.

\textbf{Data Curation Pipeline.}
We develop a data curation pipeline to generate clinically grounded doctor-patient dialogues from medical notes, consisting of three key components.
\textbf{Finding extraction:}
We extract relevant medical findings from the HPI section of discharge notes to construct patient vignettes, excluding downstream information such as lab results, treatments, and follow-up plans that would not be known during history taking.
These vignettes serve as the basis for generating patient responses and for evaluating whether a model’s question can recover the findings documented in the original note.
\textbf{Decision tree-guided dialogue generation:}
To ensure clinical relevance and alignment with diagnostic reasoning, we construct a decision tree that maps findings to candidate diagnoses. This tree provides a structured outline for guiding the LLM to generate task-oriented dialogues that reflect realistic differential diagnosis workflows.
\textbf{Critic and revision:}
While LLM-generated dialogues are generally plausible, they may omit key findings or exhibit context leakage, where the doctor infers symptoms not yet revealed by the patient. To improve quality, we introduce an LLM-based critic that identifies and corrects these issues by adding missing questions and revising premature inferences, significantly improving dialogue quality and increasing average symptom coverage.

To prepare the data for training and evaluation, we use ICD-10 codes to select discharge notes from the MIMIC-IV dataset \cite{johnson2023mimic}, focusing on two major condition groups: heart failure, cellulitis, and their associated diseases. 
This results in a diverse set of 10 clinically relevant conditions: Asthma, COPD, Cellulitis, Chronic venous insufficiency, Deep vein thrombosis, Erysipelas, Heart Failure, Necrotising Fasciitis, Pneumonia, Trauma/hematoma.
These conditions were selected based on clinical guidance to balance diagnostic challenge and feasibility, as they often present with overlapping symptoms that demand careful history taking to differentiate. 
To ensure data quality, we include only notes with a clearly defined HPI section, a primary diagnosis, and at least 100 words to guarantee sufficient clinical detail.
Using GPT-4o for data processing, our curation pipeline produces multi-turn dialogues averaging 17.8 turns for 4,972 patients across 10 diseases, which are randomly split into 4,472 for training and 500 for testing.

\textbf{Three-Stage Fine-Tuning Strategy.}
We propose a three-stage fine-tuning strategy to train LLMs for high-quality multi-turn history taking and differential diagnosis.
\textbf{Cold start with SFT:}
We initialize training using Qwen2.5-7B \cite{qwen2.5} as the base model. Using our note-guided dialogues, we apply supervised fine-tuning to teach the LLM foundational clinical reasoning and dialogue structure.
The fine-tuned model plays the doctor role and interacts with a simulated patient agent (Qwen2.5-32B), learning to follow an appropriate question-asking flow and identify relevant findings. This stage establishes a basic starting policy for structured clinical interactions.

\textbf{Self-augmentation with trajectory sampling:}
Note-guided dialogues are often overly idealized, with each doctor question reliably eliciting a relevant response, which is unrealistic in practice. 
As a result, models trained solely on these dialogues may overfit and struggle to generalize during inference. 
To improve robustness, we simulate more diverse and imperfect interactions by allowing the SFT-trained doctor model to engage in self-play with the patient agent. 
For each case, we roll out multiple dialogue trajectories and select those that achieve correct differential diagnoses with the highest recall (i.e., capturing the most documented findings). 
These selected dialogues are added to the training corpus, exposing the model to more natural conversation dynamics. This process yields 4,472 self-augmented dialogues from 67,077 successful rollouts.

\textbf{Direct preference optimization:}
While supervised fine-tuning offers a solid starting point, it does not explicitly teach the model to prefer concise, effective, and clinically sound conversations. 
To address this, we apply DPO to guide the model toward preferred history-taking behaviors. 
For each case, we generate 15 dialogue candidates per case via self-play and assign a reward score to each, based on information recall, diagnostic accuracy, and dialogue efficiency. 
Preference pairs are then constructed by contrasting top- and bottom-ranked dialogues, and the model is optimized to favor high-quality interactions. 
This step strengthens the model’s ability to ask relevant questions, avoid unnecessary turns, and make timely, well-justified diagnoses.

\textbf{Dialogue-level outcome reward:}
Designing an effective reward function is critical to the success of DPO. 
We introduce a \textit{dialogue-level} reward function that leverages the medical note as a silver-standard reference, explicitly capturing three desirable criteria: (1) thorough information gathering, (2) concise and efficient dialogue, and (3) accurate differential diagnosis.
Formally, the reward for a dialogue trajectory $\tau$ is defined as:
\begin{equation}
R(\tau) = \text{Recall} + \frac{\text{Recall}}{\text{Recall}_\text{max}} \cdot \left( 1 - \frac{\text{rank}(\text{dx}, \hat{\mathbf{y}}_T)}{K} \right) - \frac{\alpha \cdot T}{2}.
\end{equation}
Here, $\text{Recall} = \frac{|\mathcal{F}_\tau \cap \mathcal{F}|}{|\mathcal{F}|}$, measures the fraction of clinically relevant findings $\mathcal{F}$ (from the HPI) that are successfully elicited during the dialogue $\tau$, directly rewarding thorough information gathering. 
$\text{Recall}_\text{max}$ denotes the highest recall achieved across all generated dialogues.
The second term assesses diagnostic accuracy by checking whether the ground-truth diagnosis $\text{dx}$ appears within the top-$K$ predicted diagnoses $\hat{\mathbf{y}}_T$; higher rankings yield higher scores. 
In this work, we set $K = 5$ and assign $R(\tau) = 0$ when the correct diagnosis falls outside the top-$K$.
Importantly, this term is weighted by the ratio $\frac{\text{Recall}}{\text{Recall}_\text{max}}$, ensuring that high diagnostic accuracy only contributes meaningfully to the reward if sufficient relevant information has been collected, preventing reward inflation from lucky guesses. 
The final term $\frac{\alpha \cdot T}{2}$ penalizes unnecessarily long dialogues to encourage efficiency in history taking, where $T$ represents the total number of turns, and $\alpha$ is a scaling coefficient that balances the penalty relative to the other reward components.

To generate preference pairs for DPO, we roll out 15 dialogue candidates per patient through self-play and compute the mean ($\mu$) and standard deviation ($\sigma$) of their reward scores.
Dialogues with scores above $\mu + \sigma$ are labeled as high-quality, while those below $\mu - \sigma$ are considered low-quality. 
For each high-quality trajectory, we sample up to two low-quality ones to form training pairs. 
This results in 11,403 dialogue-level preference pairs used to fine-tune the model toward more informative, accurate, and efficient multi-turn clinical conversations.

\textbf{Multi-Turn History Taking via Single-Turn Reasoning.}
While multi-turn DPO aims to improve conversational models, it presents inherent limitations. 
A primary challenge is the difficulty of controlling long conversational rollouts, where early errors can compound and lead the dialogue off-track. 
The preference signal in conventional multi-turn approaches is applied to the entire trajectory, offering only coarse supervision that makes it difficult to identify and correct specific suboptimal turns. 
Furthermore, without a clear reward signal for a process like reaching a diagnosis, multi-turn models may struggle with poorly defined stopping criteria, leading to inefficient and unnecessarily long dialogues.

\textbf{Task reformulation:}
To address these challenges, we propose a novel paradigm that reframes multi-turn history taking as \textit{a sequence of single-turn reasoning problems}. Our core insight is that a multi-turn conversation can be modeled as a Markov Decision Process. By enriching the state representation at each turn to include the accumulated memory of prior turns and collected findings, we enable the model to make optimal, context-aware decisions from any intermediate step. This approach allows for stronger local supervision and improved sample efficiency, as the model's questioning strategy can dynamically adapt throughout the dialogue.
At each step, the model should perform structured reasoning through a thought process that includes \textbf{memory summarization, clinical planning, and action generation}. 
This ensures that each decision is based on the current state alone, decoupling the model's performance from the unpredictable nature of long-term conversational trajectories.

\textbf{Single-turn reasoning:}
To achieve this, we insert a structured reasoning block between each round of patient–doctor dialogue, generated by the doctor agent during inference but hidden from the patient agent.
This block, delimited by \texttt{<think>} and \texttt{</think>} tokens, comprises two key components:
Summary: A concise, structured summary of the conversation history, which is updated with each new patient response. 
This serves as the model’s working memory of the patient's condition and collected symptoms.
Planning: A clear clinical rationale for the next action. 
If the action is a follow-up question, the plan explains how it helps to confirm or rule out specific diagnoses. 
If the action is a diagnostic decision, the plan justifies the ranking of candidate diseases based on the symptoms gathered.
This structured approach enables more granular and flexible preference learning, where individual turns can be rewarded even if the full trajectory is suboptimal. 
The explicit reasoning process also enhances both interpretability and reasoning quality, as each decision is grounded in a transparent clinical rationale.
A concrete example for the proposed single-turn reasoning is shown below:

\begin{center}
\scalebox{0.9}{ 
\begin{tcolorbox}[
  colback=gray!5!white,
  colframe=black!75!black,
  title=Single-Turn Reasoning Example,
  fonttitle=\bfseries,
]
\textbf{Doctor} \\
\texttt{<think>} \\
\textbf{Summary:} Turn 0: The patient reported having a fever and shortness of breath. \\
\textbf{Plan:} I need to understand when the symptoms started and how they’ve progressed to decide what might be causing them. \\
\texttt{</think>} \\
Can you tell me when your symptoms started and how they’ve changed?


\textbf{Patient (Turn 1)} \\
The symptoms started a few days ago and have been getting worse.


\textbf{Doctor} \\
\texttt{<think>} \\
\textbf{Summary:} Turn 0: The patient reported fever and shortness of breath. \\
Turn 1: The patient said the symptoms started a few days ago and have worsened. \\
\textbf{Plan:} I need to ask more about the shortness of breath to understand its nature and possible causes. \\
\texttt{</think>} \\
Can you tell me more about the shortness of breath? Has anything helped or made it worse?
\end{tcolorbox}
}
\end{center}

\textbf{Turn-level process reward:}
The single-turn reasoning paradigm allows for a simple, interpretable reward function at each step of interaction:
\begin{equation}
R_{\text{ST}}(s_{t-1},s_t) =
\begin{cases}
\mathbb{I}[f_t \in s_t \setminus s_{t-1}], & \text{if } a_t \in \mathcal{A}^{\text{ask}} \\
\text{Recall}_t \cdot \left( 1 - \frac{\text{rank}_t}{K} \right), & \text{if } a_t \in \mathcal{A}^{\text{diagnose}} \label{eqn.st_reward}
\end{cases}
\end{equation}
Here, $\mathbb{I}[f_t \in s_t \setminus s_{t-1}]$ indicates whether a \textit{new relevant finding} was acquired at turn $t$;
$\text{Recall}_t = \frac{|\mathcal{F}_t \cap \mathcal{F}|}{|\mathcal{F}|}$ measures the proportion of ground-truth findings $\mathcal{F}$ recovered so far;
and $\text{rank}_t = \text{rank}(\text{dx}, \hat{\mathbf{y}}_t)$ gives the rank of the correct diagnosis in the top-$K$ predictions.
We assign $R_{\text{ST}}(s_{t-1},s_t) = 0$ when the correct diagnosis falls outside the top-$K$.
The agent is rewarded for acquiring new information and ranking the correct diagnosis higher, with rewards scaled by the amount of useful information gathered. 
Unlike multi-turn approaches, this setup makes question-asking and diagnosis actions directly comparable, enabling the model to learn when to stop asking and make a prediction. 
The single-turn reasoning paradigm decomposes complex dialogues into independent, context-aware decisions, allowing for precise supervision, interpretable reasoning, and verifiable rewards. 
This structure supports flexible preference learning and more accurate diagnoses, even when full dialogues are noisy or imperfect.

\textbf{Single-turn data preparation:}
Starting from our self-augmented dialogue dataset, we decompose full dialogues into individual turns and use Qwen2.5-32B to generate a structured reasoning block for each of them. 
Following the multi-turn DPO setup, we adopt a multi-stage training strategy: first fine-tuning an LLM on the single-turn augmented data, then using the model to roll out 10 candidate responses per turn by interacting with a simulated patient agent. 
After filtering out low-quality samples, we retain 80,537 context-aware single-turn interactions. 
To construct preference data for single-turn DPO, we contrast the highest- and lowest-reward responses (as defined in Eq.~\ref{eqn.st_reward}), yielding 95,811 turn-level preference pairs.

\section{Experiments}
We conduct a series of experiments to evaluate the performance of LLMs in medical history taking and differential diagnosis. 
Our evaluation begins by comparing \method{}, trained with either multi-turn (\texttt{Note2Chat-MT}) or single-turn (\texttt{Note2Chat-ST}) DPO strategies, against a diverse set of baseline LLMs. 
These include: \textit{Proprietary models:} GPT-4o \cite{achiam2023gpt}, o4-mini, Gemini-2.5 \cite{comanici2025gemini}; 
\textit{Public open-source models:} DeepSeek-R1 \cite{guo2025deepseek}, Qwen2.5 \cite{qwen2.5}, Qwen3 \cite{qwen3technicalreport};
and \textit{Domain-specific models:} HuatuoGPT-o1\cite{chen2024huatuogpt}, MedGemma \cite{sellergren2025medgemma}, DoctorAgent-RL \cite{feng2025doctoragentrl}.
We then analyze the key factors that contribute to effective history taking across clinically meaningful symptom categories.
Finally, we validate our models by comparing with practicing clinicians on a held-out test set.
Our experiments are designed to answer the following research questions:
(1) \textit{How well do existing LLMs perform in medical history taking?} 
(2) \textit{Can we improve LLMs to proactively ask follow-up questions and autonomously decide when to diagnose?}
(3) \textit{To what extent can fine-tuning and preference learning narrow the performance gap?}
(4) \textit{How does the performance of \method{} compare to that of human clinicians?}

\begin{table*}[htbp]
\centering
\small
\begin{tabular}{l|cccccc|c}
\hline
\textbf{Model} & \textbf{F1} & \textbf{Recall} & \textbf{Precision} & \textbf{Top-1} & \textbf{Top-2} & \textbf{Top-3} & \textbf{\#Turn} \\
\hline
GPT-4o & 29.2 & 33.2 & 30.5 & 49.0 & 61.4 & 67.6 & 22.9\\
o4-mini & 23.0 & 28.7 & 21.9 & 47.6 & 60.0 & 67.0 & 27.0\\
Gemini-2.5-flash & 26.6 & 35.5 & 26.7 & 51.4 & 66.2 & 73.0 & 31.9\\
\hline
Qwen2.5-7B-Instruct & 19.6 & 15.7 & 33.0 & 38.8 & 54.8 & 63.2 & 10.3\\
Qwen3-8B & 17.9 & 13.8 & 34.1  & 33.4 & 46.6 & 55.2 & 8.9\\
DeepSeek-R1-0528-Qwen3-8B & 29.6  & 34.0 & 32.7 & 37.2 & 51.6 & 61.2 & 23.4 \\
\hline
HuatuoGPT-o1-8B & 0.2 & 0.1 &1.1  & 19.4 & 33.0 & 42.8 & 2.02\\
MedGemma-4B-it & 27.2 & 31.6 & 28.0 & 40.6 & 55.2 &  62.2 & 23.4 \\
MedGemma-27B-text-it & 27.9 & 31.4 & 30.1 & 52.8 & 66.2 & 71.4 & 21.4 \\
DoctorAgent-RL & 28.4 & 35.1 & 27.5 & 35.6 & - & - &26.4 \\
\hline
\texttt{Note2Chat-MT} & \underline{43.8} & \textbf{55.4} & \underline{41.8} & \underline{62.0} & \underline{78.2} & \underline{82.6} & 27.5\\
\texttt{Note2Chat-ST} & \textbf{46.1} & \underline{46.2} & \textbf{54.5} & \textbf{70.0} & \textbf{81.2} & \textbf{84.4} & 17.3\\
\hline
\end{tabular}
\caption{History taking and diagnosis performance (\%) across different models (\textbf{Best}, \underline{Second Best})} \label{tab.main_results}
\end{table*}

\begin{table*}[htbp]
\centering
\small
\begin{tabular}{ll|cccccccc|c}
\hline
\textbf{Mode} & \textbf{Model} & \textbf{F1} & \textbf{Recall} & \textbf{Precision} & \textbf{Top-1} & \textbf{Top-2} & \textbf{Top-3} & \textbf{Avg.} &{\textbf{$\Delta$}} & \textbf{\#Turn}   \\
\hline
\multirow{4}{*}{MT} 
    & Qwen2.5-7B-Instruct & 19.6 & 15.7 & 33.0 & 38.8 & 54.8 & 63.2 & 37.5 & - &10.3\\
\hline
    & +SFT & 32.6 & 30.1 & 44.9 & 53.0 & 66.0 & 73.0 & 49.9 & +12.4 &14.1\\
    & +SFT+Self-Aug         & 40.6 & 39.2 & \underline{50.1} & \underline{62.8} & 75.0 & 81.6 & 58.2 &+20.7 &15.8\\
    & +SFT+Self-Aug+DPO     & \underline{43.8} & \textbf{55.4} & 41.8 & 62.0 & \underline{78.2} & \underline{82.6} & \underline{60.6} & \underline{+23.1} &27.5\\
\hline
\multirow{3}{*}{ST} 
    & +SFT              & 35.4 & 37.5 & 40.7 & 54.8 & 65.0 & 70.4 &50.6 &+13.1 &19.8\\
    & +SFT+Self-Aug         & 41.4 & 44.8 & 45.8 & 60.8 & 72.4 & 75.4 & 56.8&+19.3&20.5\\
    & +SFT+Self-Aug+DPO    & \textbf{46.1} & \underline{46.2} & \textbf{54.5} & \textbf{70.0} & \textbf{81.2} & \textbf{84.4} & \textbf{63.7} &\textbf{+26.2}&17.3\\
\hline
\end{tabular}
\caption{Ablation study showing the impact of each component in our \method{} framework (\textbf{Best}, \underline{Second Best}).} \label{tab.ablation}
\end{table*}

\textbf{Experiments Setup.}
Following CRAFT-MD \cite{Johri2025EvaluationFramework} and its prompting strategy, we simulate a patient agent using Qwen2.5-32B, which interacts with the evaluated LLMs acting as doctor agents. 
The doctor agent is tasked with asking relevant follow-up questions to elicit key clinical findings and terminating the conversation once sufficient information has been gathered. 
After the interview, the doctor outputs a ranked list of potential diagnoses for differential diagnosis.
To evaluate performance, we use a Qwen2.5-32B model to assess the dialogue. 
It checks how many ground-truth findings from the note are successfully elicited in the conversation and computes the rank of the true diagnosis within the predicted list. 
We report the precision, recall, F1 scores, and Top-K accuracy as our primary evaluation metrics. 
Detailed metric definitions, training setups, including hyperparameters are provided in the Appendix. 
All evaluations are conducted on our processed subset of the MIMIC-IV dataset \cite{johnson2023mimic}, with scope constrained by computational and cost considerations.

\textbf{Main Results.}
As shown in Table~\ref{tab.main_results}, \textbf{existing LLMs struggle with effective medical history taking}, which is consistent with prior studies \cite{Johri2025EvaluationFramework,Li2024MediQ}. Among them, proprietary LLMs like GPT-4o and Gemini-2.5-flash perform best, achieving top-1 diagnostic accuracies of 49.0\% and 51.4\%, respectively. In contrast, public models such as Qwen2.5-7B-Instruct and Qwen3-8B show much weaker performance, with F1 scores below 20\% and Top-1 accuracies under 40\%. DeepSeek-R1-0528-Qwen3-8B is the strongest open-source general model, reaching an F1 of 29.6\% and Top-1 accuracy of 37.2\%.
Interestingly, models like Gemini-2.5-flash and o4-mini engage in long conversations (over 27 turns on average) but still achieve low recall, indicating a lack of ability to ask clinically relevant questions. This highlights the gap between conversational fluency and clinically meaningful reasoning.

Several domain-specific medical LLMs were also evaluated. MedGemma-4B-it and MedGemma-27B-text-it perform relatively well in diagnosis (Top-1: 40.6\% and 52.8\%), likely benefiting from medical pretraining. In contrast, HuatuoGPT-o1-8B performs poorly across all metrics, especially in history taking (F1: 0.2\%), as it fails to ask follow-up questions and relies solely on the chief complaint.
DoctorAgent-RL, despite being trained on history taking dialogues, is limited by its design to predict only a single diagnosis. It performs reasonably in information elicitation (F1: 28.4\%) but fails to generalize to our setting. This is likely due to a domain mismatch, as it was trained on informal online consultations, which lack the structured, standardized symptom descriptions (e.g., onset, location, timing) found in clinical notes.

\textbf{Fine-tuning significantly improves both history taking and diagnosis.} Our proposed models, \texttt{Note2Chat-MT} and \texttt{Note2Chat-ST}, consistently outperform all baselines. Compared to the base model Qwen2.5-7B-Instruct, \texttt{Note2Chat-ST} achieves a 26.5-point absolute gain in F1 (from 19.6\% to 46.1\%), an 135.2\% relative improvement, and a 31.2-point gain in Top-1 accuracy (from 38.8\% to 70.0\%). \texttt{Note2Chat-MT} achieves the highest recall (55.4\%) but at the cost of longer dialogues (avg. 27.5 turns). In contrast, \texttt{Note2Chat-ST} delivers the best overall performance across F1, precision, and Top-K accuracy, while using fewer turns (17.3 on average). 
This demonstrates the effectiveness of the proposed \texttt{Note2Chat} framework in gathering clinically relevant information efficiently and making accurate differential diagnosis.

\textbf{Ablation Study.}
We assess the contribution of each component in our \method{} framework for medical history taking. As shown in Table~\ref{tab.ablation}, \textbf{both multi-turn (MT) and single-turn (ST) models benefit substantially from fine-tuning.} Applying SFT alone yields notable gains over the base Qwen2.5-7B-Instruct model, improving average scores by 12–13 points. However, SFT alone only makes the model competitive with top-performing LLMs, suggesting that simply memorizing dialogue patterns is insufficient for effective history taking.
Introducing self-augmented, imperfect trajectories significantly enhances recall and top-K accuracy, highlighting the value of diverse training signals. 
Further applying DPO consistently improves performance across both paradigms.
With the full pipeline (SFT + Self-Aug + DPO), \texttt{Note2Chat-MT} achieves the highest recall (55.4\%) but requires longer dialogues (27.5 turns). 
In contrast, \texttt{Note2Chat-ST} outperforms in all other metrics while using fewer turns (17.3), demonstrating the efficiency and effectiveness of the proposed single-turn reasoning framework for proactive and accurate history taking.

\begin{figure}[tbp]
  \centering
  \includegraphics[width=1\linewidth]{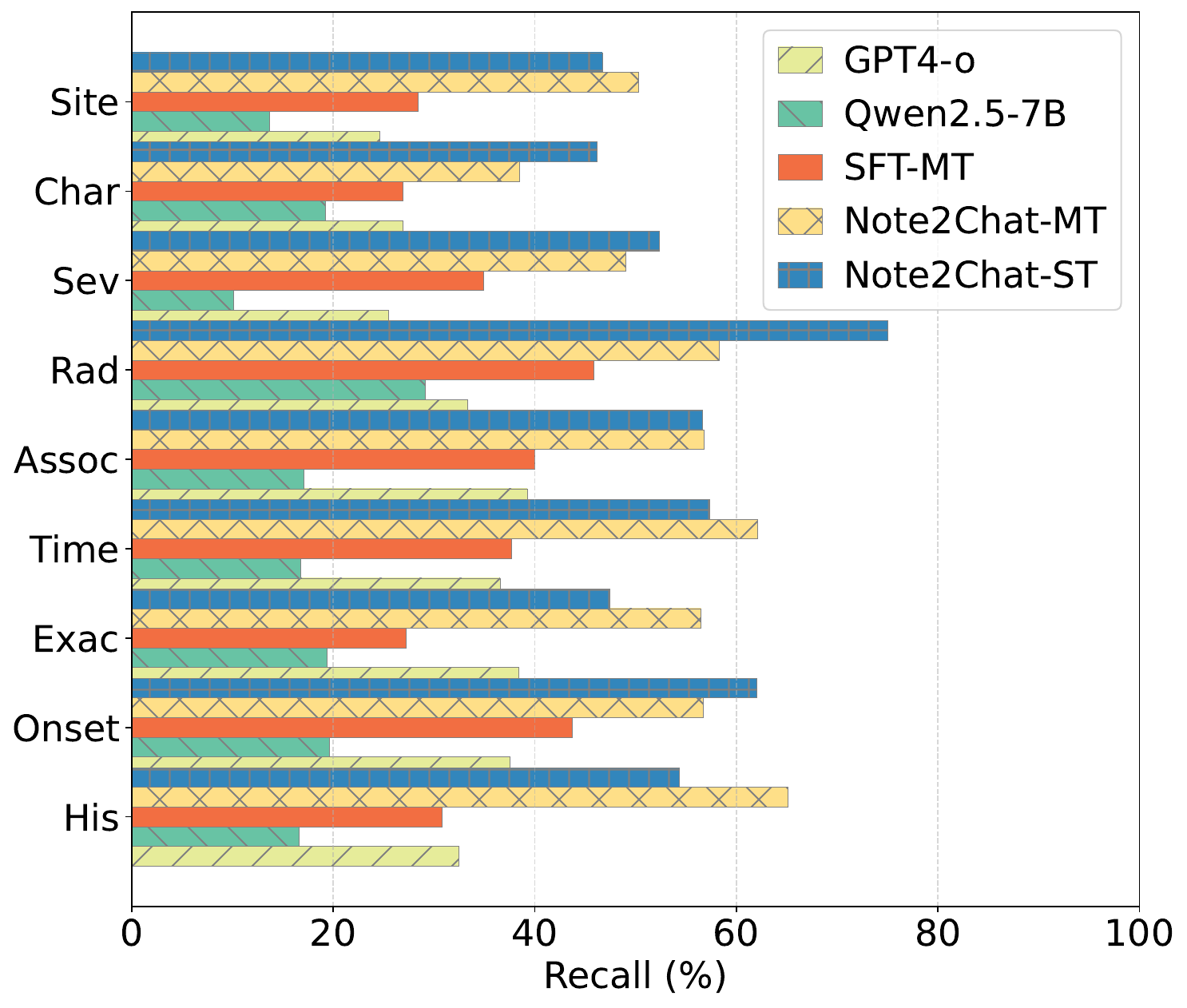}
  \caption{Recall across symptom categories.}

  \label{fig:support-category}
\end{figure}

\textbf{Analysis.}
To better understand the factors underlying effective medical history taking and how our \method{} framework improves performance, we break down recall scores by clinically meaningful symptom categories based on the SOCRATES mnemonic \cite{Mahbubani2023} (Site, Onset, Character, Radiation, Associated symptoms, Timing, Exacerbating/relieving factors, Severity) along with an additional History category. 
These categories capture essential dimensions of structured symptom characterization in clinical history taking.
As shown in Figure~\ref{fig:support-category}, even powerful proprietary models like GPT-4o underperform on essential aspects such as Site (13.6\%) and Severity (10.1\%), highlighting a gap in alignment with structured clinical inquiry. 
SFT yields moderate gains over the base Qwen2.5-7B model across all categories. 
In contrast, both \textbf{\texttt{Note2Chat-MT} and \texttt{Note2Chat-ST} achieve consistently higher recall across all categories, particularly excelling in Onset, Radiation, and History.} 
Notably, \texttt{Note2Chat-MT} reaches the highest overall recall by leveraging longer interactions (27.5 turns on average), demonstrating its effectiveness in information gathering.

\begin{figure}[tbp]
  \centering
  \includegraphics[width=0.95\linewidth]{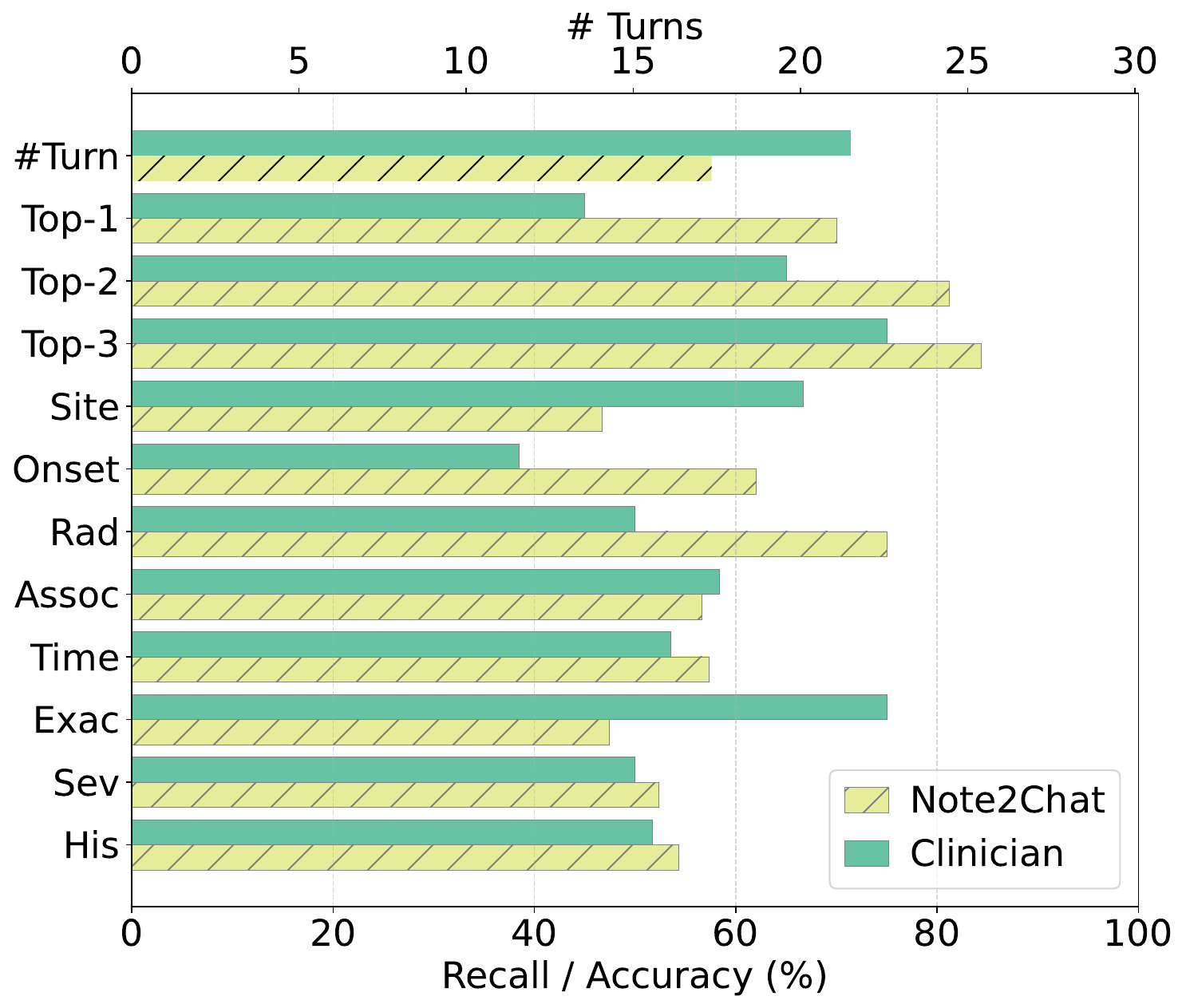}
  \caption{Comparison of history taking performance between model and clinician.}
  \label{fig:clinical_validation}
\end{figure}

\textbf{Comparison with Clinicians.}
Finally, we compare our method with a practicing clinician on a small held-out test set of 20 patient cases across 10 diseases, constrained by available resources.
As shown in Figure~\ref{fig:clinical_validation}, our model achieves comparable performance in both diagnosis accuracy and information gathering.
Notably, \textbf{it mirrors the clinician’s behavior in eliciting clinically meaningful symptoms.}
While this limited-scale validation with simulated patients is far from conclusive, it highlights the potential of LLMs for supporting real-world history taking.

\section{Conclusion}
We have presented \method{}, a note-driven framework for training LLMs to perform clinically effective history taking and differential diagnosis. By leveraging medical notes as silver-standard supervision, our approach enables models to ask relevant follow-up questions, prioritize key findings, and determine when to conclude the conversation. Through a multi-stage training pipeline combining supervised fine-tuning, self-augmented trajectory sampling, and preference optimization, \method{} achieves substantial gains in both information elicitation and diagnostic accuracy.
Our proposed single-turn reasoning paradigm enables fine-grained, verifiable supervision at each step of the dialogue. This design not only improves transparency and adaptability but also outperforms multi-turn baselines with fewer dialogue turns. Empirical results show that \method{} consistently outperforms both general-purpose and medical-domain LLMs. Additionally, clinician comparisons show promising alignment in symptom gathering, suggesting real-world applicability for AI-assisted history taking.


\section*{Acknowledgements}
This research/project is supported by the National Research Foundation Singapore under the AI Singapore Programme (AISG Award No: AISG2-TC-2023-013). This research is also supported by the SingHealth Duke-NUS Academic Medical Centre (AMC) – A*STAR Healthcare Translation Partnership (HTP) grant No. I24D1AG022 and I24D1AG085.

\newpage

\section*{Appendix}

\subsection{Evaluation Metrics}
Let $\mathcal{F} = \{f_1, \ldots, f_n\}$ be the set of \textit{ground-truth clinical findings} extracted from the note, and let $\mathcal{F}_T \subseteq \mathcal{F}$ denote the subset successfully \textit{elicited by the doctor agent} through dialogue up to turn $T$. Let $\hat{\mathbf{y}}_T = [\hat{y}_T^{(1)}, \ldots, \hat{y}_T^{(K)}] \subset \mathcal{Y}$ be the top-$K$ ranked list of diagnoses predicted at the end of the interaction, and $\text{dx} \in \mathcal{Y}$ the ground-truth diagnosis.

\begin{itemize}
\item \textbf{Precision}:

$$
\text{Precision} = \frac{|\mathcal{F}_T \cap \mathcal{F}|}{T}
$$

Measures the proportion of relevant findings elicited out of all questions asked.

\item \textbf{Recall}:

$$
\text{Recall} = \frac{|\mathcal{F}_T \cap \mathcal{F}|}{|\mathcal{F}|}
$$

Measures the proportion of ground-truth findings that were successfully elicited by the agent.

\item \textbf{F1 Score}:

$$
\text{F1} = \frac{2 \cdot \text{Precision} \cdot \text{Recall}}{\text{Precision} + \text{Recall}}
$$

Harmonic mean of precision and recall, reflecting overall elicitation effectiveness.

\item \textbf{Top-K Accuracy}:

$$
\text{Top-K Accuracy} = \mathbb{I}[\text{dx} \in \hat{\mathbf{y}}_T]
$$

Indicates whether the ground-truth diagnosis appears in the top-$K$ predictions.
\end{itemize}

\subsection{Implementation Details}
We fine-tune the Qwen2.5-7B-Instruct model in two stages: supervised fine-tuning (SFT) followed by Direct Preference Optimization (DPO). To improve training efficiency and adaptability, we apply Low-Rank Adaptation (LoRA)~\cite{hu2022lora}, using a rank of 8, scaling factor of 32, and a dropout rate of 0.05.
For SFT, the model is trained for 3 epochs with a total batch size of 128 and no gradient accumulation. 
We use a cosine learning rate scheduler with a base learning rate of $5 \times 10^{-5}$ and a warmup ratio of 0.1.
For DPO, we continue training for 2 epochs with the same batch size. 
The learning rate is set to $1 \times 10^{-5}$, also scheduled with cosine decay and a warmup ratio of 0.1.
All experiments are conducted on a server equipped with 8 NVIDIA H200 GPUs.

\begin{table}[t]
\centering
\small
\begin{tabular}{l|cc}
\hline
\textbf{LLM} & \textbf{Information Control}$\uparrow$ & \textbf{Factual Conflict}$\downarrow$ \\
\hline
Qwen2.5-7B & 81.54 & 0.37 \\
Qwen2.5-32B & 82.66 & 0.29 \\
\hline
\end{tabular}
\caption{Evaluation of the patient agent reliability.} \label{tab.patient_eval}
\end{table}

\subsection{Patient Agent Evaluation}
We evaluate the faithfulness and robustness of the simulated patient agent following protocols from prior work \cite{feng2025doctoragentrl}. Specifically, we assess: (1) \textit{Information Control}: Whether the patient only reveals information in response to doctor queries; and (2) \textit{Factual Consistency}: Whether responses align with the underlying medical note. Using GPT-4o as the evaluator, we analyze dialogues between a naive fine-tuned doctor agent and patient agents instantiated with Qwen2.5-7B and 32B. As shown in Table~\ref{tab.patient_eval}, both models perform reliably, with minimal information leakage and high factual consistency. Qwen2.5-32B shows slightly better performance and is thus selected as the patient agent in our experiments.

\subsection{Case Study}
To demonstrate the effectiveness of the proposed single-turn reasoning paradigm, we compare dialogues generated by \texttt{Note2Chat-MT} and \texttt{Note2Chat-ST} on the same patient case. Both models successfully elicit key findings and make correct diagnoses. \texttt{Note2Chat-MT} tends to ask broader questions about medical history and past treatments, which, while relevant, are more generic and less contextually grounded. In contrast, \texttt{Note2Chat-ST} better controls the conversation flow by recalling prior patient responses and asking focused, context-aware follow-up questions. Beyond improved relevance, the single-turn paradigm enhances interpretability. Each turn includes an updated memory summary and a clinical reasoning plan, offering transparency that aligns with clinical expectations and supports real-world adoption.

\begin{center}
\begin{tcolorbox}[
  enhanced,            
  colback=gray!5!white,
  colframe=black!75!black,
  title=Note2Chat-MT,
  fonttitle=\bfseries,
  breakable   
]
\textbf{Patient (Turn 0): }Hi, doctor. I’ve been having painful pressure on the right side of my chest under my right breast since yesterday, and I’ve also been feeling feverish with some really bad throat pain.\\
\textbf{Doctor: }I’m sorry you’re feeling this way. Can you tell me more about your chest pain? How does it feel exactly, and have you noticed anything else unusual along with it?\\
\textbf{Patient (Turn 1): }The chest pain is on the right side, under my breast, and it doesn't go anywhere else. It's been constant. I also have a sore throat with white spots on my tongue.\\
\textbf{Doctor: }Have you had any tests done or imaging recently related to these symptoms?\\
\textbf{Patient (Turn 2): }I had a negative rapid strep test and an unremarkable chest x-ray at the other hospital.\\
\textbf{Doctor: }Do you have any past medical conditions that might be relevant here?\\
\textbf{Patient (Turn 3): }Yes, I have primary sclerosing cholangitis that came back after a liver transplant, ulcerative colitis, and I have chronic abdominal pain managed with a narcotics contract.\\
\textbf{Doctor: }Have you been hospitalized or treated before for similar symptoms?\\
\textbf{Patient (Turn 4): }I'm not sure.\\
\textbf{Doctor: }Do you recall if you were prescribed any medications for the chest pain or throat issues?\\
\textbf{Patient (Turn 5): }I'm not sure.\\
\textbf{Doctor: }Have you had any recent trauma or injuries that might relate to your chest pain?\\
\textbf{Patient (Turn 6): }No, I haven't had any recent trauma or injuries.\\
\textbf{Doctor: }Have you noticed any changes in your bowel habits, such as diarrhea or constipation?\\
\textbf{Patient (Turn 7): }I haven't had much of a bowel movement, but I did have one loose stool earlier today in the ED.\\
\textbf{Doctor: }Aside from the chest pain and throat symptoms, have you experienced any other symptoms like fever, chills, nausea, vomiting, or difficulty breathing?\\
\textbf{Patient (Turn 8): }I've been feeling feverish and have some abdominal pain, but no nausea, vomiting, or difficulty breathing.\\
\textbf{Doctor: }Preliminary diagnoses:\\
1. Pneumonia\\
2. Gastroesophageal Reflux Disease (GERD)\\
3. Acute Coronary Syndrome\\
4. Musculoskeletal Chest Pain\\
5. Esophageal Spasm\\
\end{tcolorbox}
\end{center}

\begin{center}
\begin{tcolorbox}[
  enhanced,            
  colback=gray!5!white,
  colframe=black!75!black,
  title=Note2Chat-ST,
  fonttitle=\bfseries,
  breakable   
]
\textbf{Patient (Turn 0): }Hi, doctor. I’ve been having painful pressure on the right side of my chest under my right breast since yesterday, and I’ve also been feeling feverish with some really bad throat pain.\\
\textbf{Doctor:} \\
\texttt{<think>} \\
\textbf{Summary: }Turn 0: The patient reported experiencing painful pressure on the right side of their chest beneath the right breast since yesterday, along with symptoms of fever and severe throat pain.\\
\textbf{Plan: }I need to gather more details about the onset and progression of the chest pain. \textit{(further reasoning omitted for brevity)}\\
\texttt{</think>}\\
I’m sorry you’re feeling this way. Can you tell me a bit more about your chest pain? When did it start and has anything changed recently?\\
\textbf{Patient (Turn 1): }The chest pain started about four days ago, and it's been constant, not getting better or worse. It doesn't go anywhere else in my body.\\
\textbf{Doctor:} \\
\texttt{<think>} \\
\textbf{Summary: }Turn 0: \texttt{<Skipped. Same with the past summary.>}. \\
Turn 1: The patient has experienced constant chest pain for four days, with no significant changes in severity or spread of the pain.\\
\textbf{Plan: }I need to gather more information about the patient's overall health. \textit{(further reasoning omitted for brevity)}\\
\texttt{</think>}\\
Do you have any past medical conditions that I should know about?\\
\textbf{Patient (Turn 2): }Yes, I have primary sclerosing cholangitis, which came back after a liver transplant, ulcerative colitis, and I have chronic abdominal pain managed with a narcotics contract.\\
\textbf{Doctor:} \\
\texttt{<think>} \\
\textbf{Summary: }Turn 0–1: \texttt{<Skipped. Same with the past summary.>} \\Turn 2: The patient has primary sclerosing cholangitis that recurred post-liver transplant, ulcerative colitis, and experiences chronic abdominal pain managed under a narcotics contract.\\
\textbf{Plan: }I need to gather more details about the throat pain. \textit{(further reasoning omitted for brevity)}\\
\texttt{</think>}\\
I understand. And what about your throat? You mentioned it hurts a lot. Can you tell me more about that?\\
\textbf{Patient (Turn 3): }My throat is really sore, and I noticed white patches on my tongue.\\
\textbf{Doctor:} \\
\texttt{<think>} \\
\textbf{Summary: }Turn 0–2: \texttt{<Skipped. Same with the past summary.>} \\Turn 3: The patient reported severe soreness in the throat accompanied by the observation of white patches on the tongue.\\
\textbf{Plan: }I need to gather more information about the fever and any additional symptoms the patient might be experiencing. \textit{(further reasoning omitted for brevity)}\\
\texttt{</think>}\\
Thank you for clarifying. Besides the chest pain and throat discomfort, have you noticed any other symptoms like fever or anything else unusual?\\
\textbf{Patient (Turn 4): }I've been feeling feverish, really tired, and I have some abdominal pain that's hard to pinpoint, mostly on the right side. I also had one episode of loose stool today.\\
\textbf{Doctor:} \\
\texttt{<think>} \\
\textbf{Summary: }Turn 0–3: \texttt{<Skipped. Same with the past summary.>} \\Turn 4: The patient is experiencing chest pain, throat discomfort, fever, fatigue, right-sided abdominal pain, and one episode of loose stool.\\
\textbf{Plan: } Asking about changes in bowel habits or difficulty passing stools can help identify if there's an issue with the digestive tract that might be contributing to these symptoms. \textit{(further reasoning omitted for brevity)}\\
\texttt{</think>}\\
Have you noticed any changes in your bowel habits or had difficulty passing stools recently?\\
\textbf{Patient (Turn 5): }I haven't had any problems passing stools, but I did have one episode of loose stool today.\\
\textbf{Doctor:} \\
\texttt{<think>} \\
\textbf{Summary: }Turn 0–4: \texttt{<Skipped. Same with the past summary.>} \\Turn 5: The patient reported no difficulties in passing stools regularly but noted one episode of loose stool on the day of the consultation.\\
\textbf{Plan: }I need to gather more information about the patient's current medication regimen. \textit{(further reasoning omitted for brevity)}\\
\texttt{</think>}\\
Thank you for sharing that. Could you tell me more about any medications or treatments you’re currently taking?\\
\textbf{Patient (Turn 6): }I'm on a narcotics contract for managing my chronic abdominal pain.\\
\textbf{Doctor:} \\
\texttt{<think>} \\
\textbf{Summary: }Turn 0–5: \texttt{<Skipped. Same with the past summary.>} \\Turn 6:The patient is currently on a narcotics contract for managing their chronic abdominal pain.\\
\textbf{Plan: }I need to inquire about any recent changes in the patient's bowel habits or the presence of blood. \textit{(further reasoning omitted for brevity)}\\
\texttt{</think>}\\
Have you noticed any changes in your bowel habits recently, or have you seen any blood in your stool?\\
\textbf{Patient (Turn 7): }I haven't seen any blood in my stool, but I did have one episode of loose stool today.\\
\textbf{Doctor:} \\
\texttt{<think>} \\
\textbf{Summary: }Turn 0–6: \texttt{<Skipped. Same with the past summary.>} \\Turn 7: The patient reported having one episode of loose stool but did not observe any blood in the stool.\\
\textbf{Plan: }I need to clarify if there have been any changes in the patient's appetite or weight. \textit{(further reasoning omitted for brevity)}\\
\texttt{</think>}\\
Have you had any changes in your appetite or weight recently?\\
\textbf{Patient (Turn 8): }I haven't eaten much in at least two days.\\
\textbf{Doctor:} \\
\texttt{<think>} \\
\textbf{Summary: }Turn 0–7: \texttt{<Skipped. Same with the past summary.>} \\Turn 8: The patient has not consumed food in at least two days, indicating a recent change in their appetite.\\
\textbf{Plan: }Given the patient's symptoms and medical history, I'm considering several preliminary diagnoses: Pneumonia, Acute Coronary Syndrome, Gastroenteritis, Non-specific Viral Syndrome,Chronic Pain Flare-Up.\\
\texttt{</think>}\\
Preliminary Diagnoses:\\
1. Pneumonia\\
2. Acute Coronary Syndrome\\
3. Gastroenteritis\\
4. Non-specific Viral Syndrome\\
5. Chronic Pain Flare-Up\\

\end{tcolorbox}
\end{center}

\subsection{Prompt Details}
In this section, we provide the detailed prompts used for data curation, doctor and patient agent construction, single-turn reasoning generation, and history taking evaluation.

\subsubsection{Prompts for History Taking Data Curation}

\begin{center}
\begin{tcolorbox}[
  enhanced,            
  colback=gray!5!white,
  colframe=black!75!black,
  title=Decision Tree Generation,
  fonttitle=\bfseries,
  breakable   
]
 Based on the provided medical note below, generate a structured decision tree for differential diagnosis. Terminate branches with potential diagnoses, including the final diagnosis as the confirmed condition. Prioritize clinical relevance and logical progression. Ensure the decision tree incorporates all symptoms mentioned in the history of present illness (HPI) to the greatest extent possible.\\
Structure the output in JSON format with the following keys:\\
tree: A nested object where each node contains:\\
criteria: Short clinical question or finding (e.g., 'Fever present?').\\
branches: Sub-nodes for 'yes'/'no' responses (if applicable).\\
diagnoses: List of potential diagnoses (if terminal node), each with:\\
condition: Diagnosis name.\\
confidence: Likelihood (e.g., 'high', 'moderate', 'low').\\
is\_final: Boolean indicating if it matches the final diagnosis.\\

Medical Note:\\
Chief Complaint: \{chief\_complaint\}\\
HPI: \{hpi\}\\
Final Diagnosis: \{diagnosis\}\\

Provide only the JSON output, without additional text.
\end{tcolorbox}
\end{center}

\begin{center}
\begin{tcolorbox}[
  enhanced,            
  colback=gray!5!white,
  colframe=black!75!black,
  title=Dialogue Generation,
  fonttitle=\bfseries,
  breakable   
]
Generate a history-taking dialogue between a doctor and a patient based on the decision tree and the medical note.  
The conversation opens with a natural and concise initial statement from the patient, describing clear symptoms from Patient's chief complaint or only one symptom History of Present Illness, and concludes with five preliminary differential diagnoses—ranked by clinical likelihood—provided by the doctor.

To make the conversation clinically convincing, the doctor knows nothing about the patient in advance and the doctor should only learn information through the patient's responses, ask appropriate follow-up questions in plain, patient-friendly language, and actively compare, rule in, and rule out the potential diseases without using unexplained medical jargon.

The patient should answer exclusively based on the provided medical note, using layperson terms. Avoid volunteering unsolicited information; answers should be direct and relevant to the doctor’s questions, and strictly adhere to the facts in the medical note without fabricating or altering any information.

\textbf{Input:}

Decision Tree: \{decision\_tree\}

Medical Note:

Chief Complaint: \{chief\_complaint\}

HPI: \{hpi\}

Final Diagnosis: \{diagnosis\}
\begin{verbatim}
{
  "conversation": [
    {
        "role": "patient", 
        "content": "..."
    },
    ...
    {
        "role": "doctor", 
        "content": "diagnosis..."
    }
  ],
  "preliminary_diagnosis": [
    {
      "disease": "Disease A",
      "reason": "Explanation"
    },
    {
      "disease": "Disease B",
      "reason": "Explanation"
    }
  ]
}
\end{verbatim}

\end{tcolorbox}
\end{center}

\begin{center}
\begin{tcolorbox}[
  enhanced,            
  colback=gray!5!white,
  colframe=black!75!black,
  title=Dialogue Revision,
  fonttitle=\bfseries,
  breakable   
]
 You are criticizing a dialogue from an AI doctor agent asking a patient about their symptoms over an online chat interface (because it is virtual, the clinician cannot do physical exams like in a clinic). The patient is suffering from a particular medical problem, and the doctor hopes to understand their chief complaint, history of present illness in order to best determine what is going on and their likely diagnoses. The doctor must dig deep(potentially asking followup questions) into the particular symptoms the patient is complaining about and anything clinically significant. \\

Make suggestions for the doctor to better meet the following criteria:\\
- The doctor agent avoids asking too many questions, focusing on a maximum of one or two per response.\\
- The responses should not reveal that the doctor agent is an AI chatbot. They should flow naturally, maintain factual accuracy, and facilitate further engagement from the patient.\\
- The doctor should only learn medical information from what the patient says during the conversation. They should not reference any lab results, diagnoses, hospitalizations, or medications unless the patient has brought them up.\\

Make suggestions for the patient to better meet the following criteria:\\
- The patient may respond only with facts from the Medical Note—no guessing or assumptions—using simple, layman-friendly language without medical jargon.\\

Your subtasks: \\
1. Missing facts:\\
For each item in missing\_facts, assume that the doctor has no prior knowledge of this information---not even a hint. The doctor must not mention or imply any part of the fact directly. Instead, the doctor should ask a natural, general, and open-ended question to give the patient an opportunity to bring up the information themselves.
If the missing fact relates to past medical history, the doctor should ask a single, general question (e.g., “Do you have any past medical conditions?”), and the patient should respond using simple, layman-friendly language without going into overly specific medical details. Demographic information can be ignored.\\

2. Logical inconsistencies  \\
For each turn in the conversation, detect instances where the doctor references facts that the patient has not mentioned, or where the patient volunteers unsolicited information. Fix these by using either an add\_turn (to insert a new turn after a specific one) or a revise\_turn (to replace the problematic turn). Make sure to evaluate every turn in the dialogue for these issues.\\

Input (provided to you)\\
Medical Note: Chief Complaint: \{chief\_complaint\} HPI: \{hpi\} Final Diagnosis: \{diagnosis\} Missing Facts (not mentioned in the conversation): \{missing\_facts\} Conversation: {conversation} \\

Please respond strictly in the following JSON format and do not include any other text: \\
\begin{verbatim}
{
  "critic_res": [
    {
      "action": "add_turn" or 
        "revise_turn",
      "location": <even turn number>,
      "doctor": "<revised or new 
        doctor utterance>",
        only for add_turn
      "patient":  "<corresponding 
        patient reply>", //only for 
        add_turn
      "comment":  "Why this exchange 
      is needed."
    }
  ]
}
\end{verbatim}

\end{tcolorbox}
\end{center}

\subsubsection{Prompts for the Doctor Agent}
\begin{center}
\begin{tcolorbox}[
  enhanced,            
  colback=gray!5!white,
  colframe=black!75!black,
  title=Doctor Prompt: For Fine-Tuned Models,
  fonttitle=\bfseries,
  breakable   
]
      You are a doctor. \\
      Your task is to gather the patient's symptoms without giving explanations or sharing impressions, simulating real-world medical interviews. In conducting the history-taking, you should use a
decision tree framework for differential diagnosis. To make the conversation 
clinically convincing, you should ask appropriate follow-up questions in plain, patient-friendly language, 
and actively compare, rule in, and rule out the potential diseases without using unexplained medical jargon. \\

For each symptom that arises, consider—when clinically relevant—asking about: \\
1. Site – exact location \\
2. Onset – when it began (sudden vs. gradual) \\
3. Character – quality or nature (e.g., sharp, dull, burning) \\
4. Radiation – whether it spreads elsewhere \\
5. Associated symptoms – what occurs alongside it \\
6. Timing – duration, frequency, pattern\\
7. Exacerbating / Relieving factors – what worsens or improves it\\
8. Severity – intensity on an easy-to-understand scale\\

After gathering sufficient information, state five final diagnoses.
\end{tcolorbox}
\end{center}

\begin{center}
\begin{tcolorbox}[
  enhanced,            
  colback=gray!5!white,
  colframe=black!75!black,
  title=Doctor Prompt: For Pre-Trained Models,
  fonttitle=\bfseries,
  breakable   
]
You are an AI doctor. Arrive at a diagnosis of a patient's medical condition. \\
Ask only one question at a time, and it should not be more than 1 line. \\
Continue asking questions until you're 100\% confident of the diagnosis. \\
Do not ask the same question multiple times. \\
Ask different questions to cover more information. \\
The questions should cover current symptoms, medical history of illness and medications, and relevant family history if necessary. \
Keep your questions short and brief to not confuse the patient. \\
After you're done asking questions, give the preliminary diagnosis as a short response.\\
You must state **preliminary diagnoses:** at the beginning of your response, otherwise you will be penalized.\\
You must give five diagnoses otherwise you will be penalized.

\end{tcolorbox}
\end{center}

\begin{center}
\begin{tcolorbox}[
  enhanced,            
  colback=gray!5!white,
  colframe=black!75!black,
  title=Doctor Prompt: For Differential Diagnoses,
  fonttitle=\bfseries,
  breakable   
]
You are an AI doctor. Based on the patient's answers so far, generate a list of exactly five possible diagnoses.
Only output the final diagnoses.
Do not ask any further questions.
Begin your response with 'preliminary diagnoses:' otherwise you will be penalized.
List exactly five potential diagnoses based on the information available.
Be concise and medically accurate.
\end{tcolorbox}
\end{center}

\subsubsection{Prompt for the Patient Agent}

\begin{center}
\begin{tcolorbox}[
  enhanced,            
  colback=gray!5!white,
  colframe=black!75!black,
  title=Patient Prompt: Vignette-Guided Responses,
  fonttitle=\bfseries,
  breakable   
]
You are a patient. \\
You do not have any medical knowledge. \\
You have to describe your symptoms from the given case vignette based on the questions asked. \\
If the information is not mentioned in the given case vignette, respond with exactly "I don't know." \\
If the question is repetitive, respond with exactly "Sorry, you've already asked this question." \\
Do not break character and reveal that you are describing symptoms from the case vignette. \\
Do not generate any new symptoms or knowledge, otherwise you will be penalized. \\
Do not reveal more information than what the question asks. \\
Keep your answer short, to only 1 sentence. \\
Simplify terminology used in the given paragraph to layman language. \\
**Case Vignette**: \{case\_desc\}
\end{tcolorbox}
\end{center}

\subsubsection{Prompts for Single-Turn Reasoning Generation}

\begin{center}
\begin{tcolorbox}[
  enhanced,            
  colback=gray!5!white,
  colframe=black!75!black,
  title=History Summarization,
  fonttitle=\bfseries,
  breakable   
]

Rewrite the following doctor-patient conversation into a third-person, concise statement describing the patient's condition.\\
Accurately describe the situation without omitting any details or adding any information not present in the original conversation.\\
Conversation:\\
\{conversation\}
\end{tcolorbox}
\end{center}

\begin{center}
\begin{tcolorbox}[
  enhanced,             
  colback=gray!5!white,
  colframe=black!75!black,
  title=Clinical Plan Generation,
  fonttitle=\bfseries,
  breakable   
]
Based on the current conversation summary and the doctor’s next action, write the doctor’s internal reasoning from a first-person perspective.\\
The explanation should sound natural and concise, focusing on the medical diagnostic rationale—specifically, which potential conditions this action could help rule in or rule out.\\
conversation summary: \{conversation\_summary\}\\
next action: \{next\_action\}\\
Your internal reasoning:
\end{tcolorbox}
\end{center}

\begin{center}
\begin{tcolorbox}[
  enhanced,            
  colback=gray!5!white,
  colframe=black!75!black,
  title=Differential Diagnosis Reasoning,
  fonttitle=\bfseries,
  breakable   
]

Based on the current conversation summary and the doctor’s next action (preliminary diagnoses), write the doctor’s internal reasoning from a first-person perspective.\\
The explanation should be natural and concise, clearly state that these are preliminary diagnoses, and explain the medical diagnostic rationale—why these diseases are being considered and why some are more likely than others at this stage.\\
conversation summary: \{conversation\_summary\}\\
next action: \{next\_action\}\\
Your internal reasoning:
\end{tcolorbox}
\end{center}

\subsubsection{Prompts for History Taking Evaluation}
\begin{center}
\begin{tcolorbox}[
  enhanced,            
  colback=gray!5!white,
  colframe=black!75!black,
  title=Medical Finding Checking,
  fonttitle=\bfseries,
  breakable   
]
You are given two inputs:\\
A list of sentences from a patient's medical note.\\
A multi-turn conversation between a patient and a doctor.\\

Your goal is:\\
For each sentence in the medical note:\\
Identify the latest turn in the conversation (whether spoken by the doctor or the patient) that mentions the information in that sentence, based on meaning.\\
If a matching statement appears, output the turn number where it first appears.\\
If no statement in the conversation corresponds to the sentence, output -1.\\

Important rules:\\
Match based on meaning, not necessarily exact wording.\\
Consider both doctor and patient utterances.
Pick the latest (last) turn that matches.
Be strict: if the information was not mentioned, output -1.\\

Input:\\
sentences of medical note:
\{note\_sentences\}\\
conversation: 
\{conversation\}\\
Please respond strictly in the following JSON format and do not include any other text:\\
\begin{verbatim}
[
  {
    "index": 0,
    "sentence": "<medical note 
        sentence>",
    "turn": <turn number> // use -1 
        if not asked
  },
  ...
]
\end{verbatim}

\end{tcolorbox}
\end{center}

\begin{center}
\begin{tcolorbox}[
  enhanced,            
  colback=gray!5!white,
  colframe=black!75!black,
  title=Diagnoses Evaluation,
  fonttitle=\bfseries,
  breakable   
]

You are given a ground truth diagnosis and a list of candidate diseases.\\
Your task is to determine the index (starting from 0) of the first disease in the list that is a valid match for the ground truth diagnosis based on **medical meaning**.\\
A candidate disease is considered a match if:\\
- It exactly matches the ground truth diagnosis, OR \\
- It is a **more specific subtype** of the ground truth diagnosis — that is, the ground truth is a **broader category** that includes the candidate disease.\\
Do not match based on text similarity alone. Use your medical knowledge to judge whether the candidate disease is a specific instance of the broader ground truth diagnosis.\\
ground truth diagnosis: \{diagnosis\}\\
candidate diseases: \{ddx\_list\}\\
Return your result in the following JSON format:\\
\begin{verbatim}
{
  "match_index": INDEX
}
\end{verbatim}

\end{tcolorbox}
\end{center}

\bibliography{aaai2026}

\end{document}